\documentclass[lettersize,journal]{IEEEtran}
\usepackage{amsmath,amsfonts}
\usepackage{algorithmic}
\usepackage{algorithm}
\usepackage{array}
\usepackage[caption=false,font=normalsize,labelfont=sf,textfont=sf]{subfig}
\usepackage{textcomp}
\usepackage{stfloats}
\usepackage{url}
\usepackage{verbatim}
\usepackage{graphicx}
\usepackage{cite}
\usepackage{amsmath}
\usepackage[normalem]{ulem}
\useunder{\uline}{\ul}{}
\usepackage{adjustbox}
\usepackage{enumitem}
\usepackage[usenames,dvipsnames]{color}
\usepackage{ulem}
\usepackage[numbers,sort&compress]{natbib}

\begin{document}

\title{Depth Estimation from a Single Optical Encoded Image using a Learned Colored-Coded Aperture}

\author{Jhon Lopez, Edwin Vargas, ~\IEEEmembership{Student Member,~IEEE,} Henry Arguello, ~\IEEEmembership{Senior Member,~IEEE,}
        % <-this % stops a space
\thanks{This work was supported by the Vicerrectoría de Investigación y Extensión of the Universidad Industrial de Santander under the research project 3735 titled "Optical-computer system for acquisition of encoded images for the preservation of privacy and actions recognition in clinical environments"}}% <-this % stops a space
%\thanks{Manuscript received April 19, 2021; revised August 16, 2021.}}



\maketitle

\begin{abstract}
Depth estimation from a single image of a conventional camera is a challenging task since depth cues are lost during the acquisition process. State-of-the-art approaches improve the discrimination between different depths by introducing a binary-coded aperture (CA) in the lens aperture that generates different coded blur patterns at different depths. Color-coded apertures (CCA) can also produce color misalignment in the captured image which can be utilized to estimate disparity.  Leveraging advances in deep learning, more recent works have explored the data-driven design of a diffractive optical element (DOE) for encoding depth information through chromatic aberrations. However, compared with binary CA or CCA, DOEs are more expensive to fabricate and require high-precision devices. Different from previous CCA-based approaches that employ few basic colors, in this work we propose a CCA with a greater number of color filters and richer spectral information to optically encode relevant depth information in a single snapshot. Furthermore, we propose to jointly learn the color-coded aperture (CCA) pattern and a convolutional neural network (CNN) to retrieve depth information by using an end-to-end optimization approach. We demonstrate through different experiments on three different data sets that the designed color-encoding has the potential to remove depth ambiguities and provides better depth estimates compared to state-of-the-art approaches. Additionally, we build a low-cost prototype of our CCA using a photographic film and validate the proposed approach in real scenarios.
\end{abstract}


\section{Introduction}
\IEEEPARstart{W}{hen} a scene is captured with a conventional camera, the three-dimensional information of the scene is lost and only a two-dimensional representation of the scene is obtained. The retrieval of 3D scene information, particularly the construction of a depth map, has a wide range of applications, including autonomous navigation \cite{royo2019overview}, industrial manufacturing \cite{ni2016application}, 3D object detection \cite{shrivastava2013building}, among others. State-of-the-art approaches for depth estimation can be broadly divided into active and passive techniques.  Active techniques like LiDAR and structured light imaging typically employ light sources and specialized time-of-flight (ToF) scanning systems \cite{wang2020review}. Passive approaches, as the multi-camera system \cite{liu2009continuous} perform triangulation between corresponding points on multiple views of the scene for depth estimation \cite{andrew2001multiple}.  Stereoscopic and multi-view methods are quite accurate in estimating depth, and have been applied in a wide variety of applications as 3D scene reconstruction \cite{liu2009continuous, hane2013joint}. However, there is a continuing need for compact and low-consumption passive depth estimation technology that is not satisfied with multiple cameras on a single platform. Additionally, while active depth estimation technologies such as LiDAR can provide accurate depth estimation, they require extra components and great energy demand, which can make their usage impractical for certain applications.

On the other hand, monocular vision approaches aim to estimate depth maps from a single image. Traditional approaches of this type are based on Markov Random Fields \cite{saxena2008make3d}, boosted classifiers \cite{ladicky2014pulling}, or depth from defocus (DFD). DFD approaches recover the depth information from two differently-focused scene images, and it has been studied extensively in computer vision for almost three decades \cite{favaro2010recovering, alexander2016focal, tang2017depth,carvalho2018deep}. Although the basic theory behind this technique is well known, DFD has found limited use in practice because it requires static scenes, dense surface texture, and images with significant defocus blur \cite{tang2017depth}. Related approaches, based on defocus, design an optically controlled blur that varies with the scene depth to encode the depth information in a single image using coded apertures (CAs) \cite{levin2007image, shedligeri2017data}. For example, Levin et al. \cite{levin2007image} design a binary CA to generate optically depth-dependent blur that enables the estimation of an all-in-focus image and a depth map from the acquired coded image. Bando et al. \cite{bando2008extracting} proposed a novel method for acquiring three shifted views of a scene based on optical geometry using a color-coded aperture (CCA). This technique enables the acquisition of multiple views of a scene in the RGB planes of a single exposure, resulting in color misalignment that is depth-dependent. In other words, the captured image provides depth information by analyzing the color shifts induced by the CCA.
 
On another front, the large amount of available labeled data has led to the rapid growth of deep learning and the development of deep neural network (DNN) architectures with outstanding performance on different areas \cite{ronneberger2015u,wang2016cost}. For depth estimation, developments in convolutional neural networks (CNNs) have shown that pixel-level depth maps can be recovered from a single image \cite{carvalho2018deep,zhao2020monocular}. More recently, a new paradigm, called deep optics \cite{sitzmann2018end,arguello2023deep}, exploits deep learning techniques to develop cameras for specific purposes, such as depth estimation \cite{chang2019deep,wu2019phasecam3d}, spectral imaging  \cite{baek2021single, vargas2021time}, light field \cite{wang2018end}, super-resolution \cite{sitzmann2018end}, among others. In particular, applications of deep optics for depth estimation include learning a diffractive optical element (DOE) to promote chromatic aberrations as an additional depth-cue to be decoded by a neural network \cite{chang2019deep,wu2019phasecam3d,ikoma2021depth}. Although the design and implementation of DOEs have demonstrated high precision in depth estimation, the fabrication of DOEs is expensive, requires high-precision devices, and can be easily scratched or damaged.% \jhon{making them impractical, thus the need to explore other elements, which encode accurate depth information, such as coded apertures.}

In this work, we propose to design a low-cost and readily manufacturable CCA to optically encode relevant depth information in a single snapshot.  While previous research has primarily focused on optical coding with a limited set of basic colors, we propose a CCA with a greater number of color filters and richer spectral information. We leverage the advances of deep learning techniques to jointly learn the optical modulation and the computational decoding algorithm in an end-to-end (E2E) framework. We demonstrate through various computational simulations that our encoding approach outperforms other state-of-the-art encoding alternatives based on CAs, CCAs, or DOEs. Furthermore, we build a prototype of the designed CCA using a photographic film and estimate high-fidelity depth maps demonstrating a cost-effective and practical solution.

\section{Related Work}

The concept of depth estimation refers to the process of preserving the 3D information of the scene using the 2D information captured by the camera sensors. This task of preserving 3D scene information is currently an area of study in optics and computer vision. Traditional methods for image acquisition and depth estimation can be classified into two sets, active and passive methods.

\subsection{Active Methods}

Active methods measure the distance to the objects by measuring the time elapsed from sending pulse signals and their bounce back. These methods are generally based on computing the time of flight of laser, ultrasound, or radio signals of the electromagnetic spectrum to measure and search for objects \cite{salman2017distance}. Usually, ToF sensors are more suitable for indoor scenarios and short-range $(<2[~m])$ depth sensing, and laser-based scanners (LiDAR) and radars are commonly utilized for 3D measurement outdoors. Key advantages of LiDAR sensors include high resolution, accuracy, high performance in low light, and speed. However, LiDARs are expensive, requiring extensive power resources, thus making them unsuitable for consumer products.

\subsection{Passive Methods}

Passive methods exploit characteristics of the captured images, such as the relative size of objects, textures, and blur, among others to estimate depth. In this category of passive methods, there are two main approaches: multi-view depth estimation, such as depth from stereoscopic images, and monocular depth estimation. Stereo imaging is a typical, passive, frequently used depth technique, however, it can present ambiguity in texture information, so the only solution to this problem is to use more cameras \cite{okutomi1991multiple}, which increases the cost, space of these systems, and calibration.  An alternative approach to these drawbacks is to use a single-camera (monocular) system, which is fundamentally different from depth estimation from stereo pairs because triangulation is no longer possible. In a monocular setting, contextual information is required, e.g., texture variation, occlusions, scene context \cite{khan2020deep}, but mainly over the years, defocus-blur has been explored as a depth cue.

\subsection{Coded Defocus Blur}
% \textcolor{blue}{
%Optical blur is typically described as a convolution of the original sharp scene with a blur kernel known as the point spread function (PSF). In order to recover the original image, deconvolution methods must be used; however, traditional blur kernels are circular and promote noise, making the reconstruction of the latent image difficult \cite{takeda2012coded}. Levin et al. \cite{levin2007image} attempted to estimate a sharp image by using a binary CA and generating an optimized blur pattern. Furthermore, because the size of the defocus kernel changes in terms of the distance between the camera and the object, knowledge of the object's depth can be estimated. %\EVcomment{for?. I never read this section. It is very confusing}. \jhon{This is where the concept of DFD comes into play, which involves depth estimation by exploiting the phenomenon of optical defocus. Several studies have exploited this technique, including Schechner et al. \cite{schechner2000depth}.}

%Many experiments have been conducted to make depth from defocus robust by calculating depth from several photographs recorded with varied optical conditions. 
Depth from defocus (DFD) can be achieved by exploiting the fact that optical blur varies with the scene depth. Specifically, DFD approaches recover depth information from two differently-focused scene images, and it has been studied extensively in computer vision for almost three decades \cite{favaro2010recovering, alexander2016focal, tang2017depth,carvalho2018deep}. 
Nayar et al. \cite{nayar1996real} proposed to use a prism between a lens and an image sensor to capture two photos with varying focus distances from the same viewpoint. The authors in \cite{hiura1998depth} employed a multi-focus camera to collect three photos with varying focus distances, and introduced a coded aperture to estimate the depth map and a blur-free image of the scene using an inverse filter \cite{takeda2012coded}. However, multiple captures with coded apertures are only feasible for static scenes. 

To overcome this limitation, various approaches harness the advantages of CCAs \cite{amari1992single, chakrabarti2012depth, bando2008extracting, paramonov2016depth}. Optically, CCAs can induce different degrees of blur in each RGB channel of the captured image generating color misalignment along channels. Therefore, depth estimation can be inferred from the disparity observed across the sensor channels eliminating the need for multiple captures.
More precisely, the use of CCA for depth estimation was first proposed in 1993 in \cite{amari1992single}, by using red and green filters jointly with an algorithm based on the color gradient, to estimate the disparity. Bando et al. \cite{bando2008extracting} also explored a CCA by dividing the aperture into three regions through which light can only pass in one of the RGB color bands, acquiring three displaced views of a scene in the RGB planes. Bando et al. \cite{bando2008extracting} proposed to include the blue color and divide the CCA into three regions in order to acquire three displaced views of a scene in the RGB planes. Since \cite{amari1992single} and \cite{bando2008extracting} have low light efficiency, a high transmittance CCA has been proposed by using the CMYK color space \cite{paramonov2016depth}. Fig.\ref{fig:prev_cca} shows state-of-the-art  CCAs for depth estimation.

\subsection{Deep Optics}

Optics and algorithm design are both involved in computational imaging systems. Deep optics treats the entire system as one neural network and develops an end-to-end optimization framework, rather than optimizing these two components separately and sequentially. The first layer of the network, in particular, corresponds to physical optical elements, while all subsequent layers represent the computational algorithm. All the parameters are learned over a large dataset using a task-specific loss. This technique has been successfully applied in specific tasks such as spectral imaging \cite{arguello2021shift, vargas2021time}, depth estimation \cite{wu2019phasecam3d, chang2019deep, haim2018depth}, light field \cite{wang2018end}, super-resolution imaging \cite{sitzmann2018end} and in compound task as spectral-depth reconstruction \cite{baek2021single}. 

\begin{figure}[bt]
    \centering
    \includegraphics[width =\columnwidth]{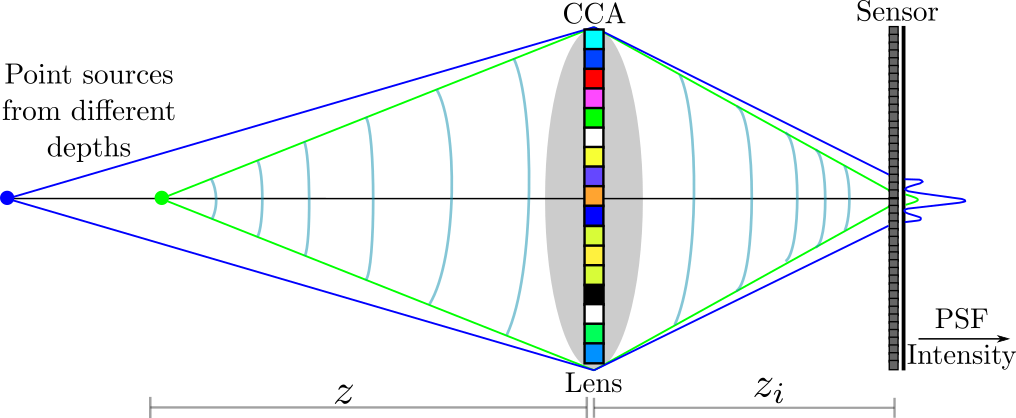}
    \caption{Optical propagation model of point sources through a color-coded aperture in front of a thin lens.} %\EVcomment{Improve resolution and colors. Would be nice to improve the CA!!}
    \label{fig:propagation}
\end{figure}
\section{Proposed Method} \label{sec:proposed}

\begin{figure*}[t]
    \centering
    \includegraphics[width=\textwidth]{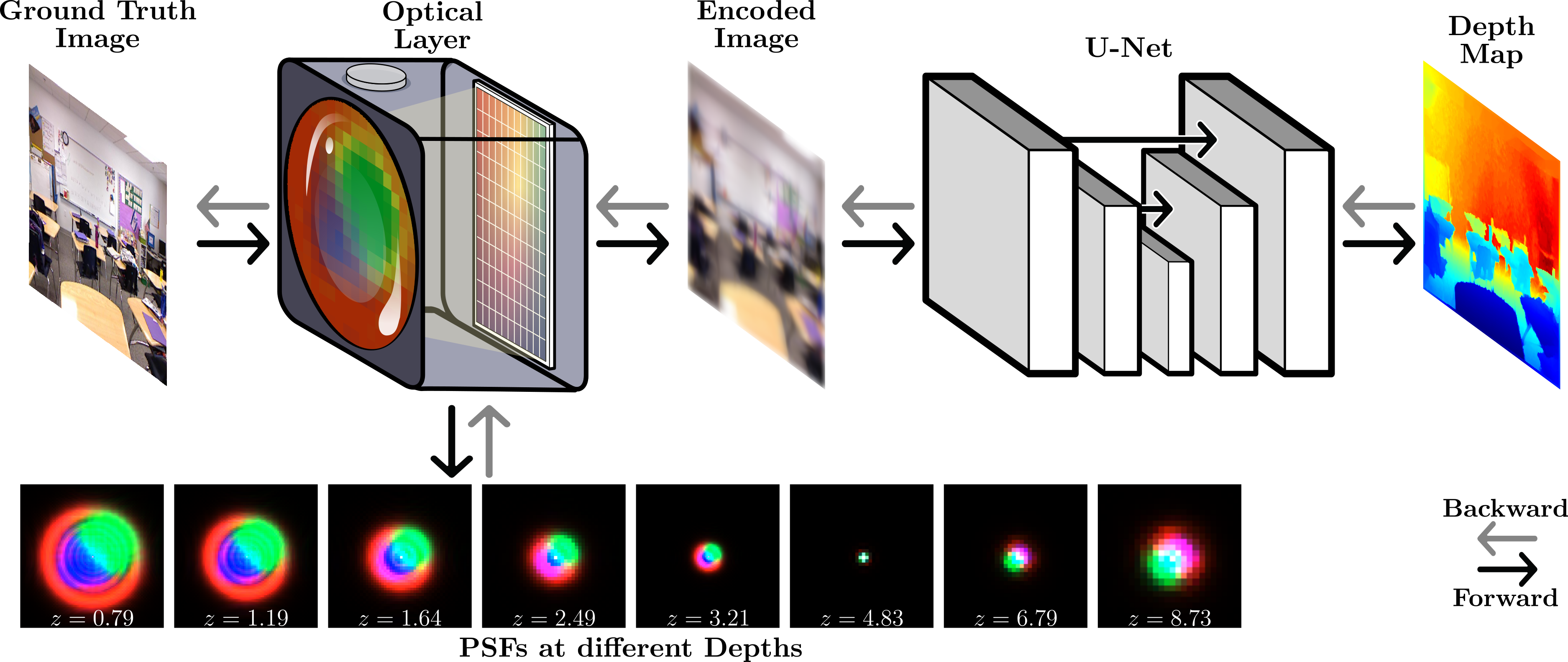}
    \caption{
    %\EVcomment{This image looks out of proportion. Could you reduce de size of UNET and images a little bit. The PSFs look saturated. I already told you to replace with PSFs in Fig 9.} 
    Our proposed architecture consists of two parts. In the optical layer, digital modeling of depth-dependent PSFs is performed from an optimized color aperture code, thus obtaining a depth-encoded image. In the reconstruction network, we use a U-Net-based network for depth estimation from the coded image. The optical layer and reconstruction network parameters are optimized based on the defined loss between the estimated depth and the reference depth map.}
    \label{fig:method}
\end{figure*}

This work considers a monocular system that encodes depth information by leveraging a CCA, as shown in Fig. \ref{fig:propagation}. In contrast with state-of-the-art CCA with few basic colors as shown in Fig. \ref{fig:prev_cca}, we propose to design CCA with a greater number of color filters and richer spectral information. As we mentioned before, the performance of this system mainly depends on the CCA and the recovery algorithm used to decode the depth information. Therefore, we propose to jointly learn the CCA of the optical encoder and a neural network working as a decoder in an end-to-end manner. Fig. \ref{fig:method} presents our proposed solution consisting of two main components: 1) a differentiable optical layer, whose trainable parameters are the color-coded aperture values, and 2) a U-Net network \cite{ronneberger2015u} for depth estimation. During training, the optical layer is fed with a given all-in-focus spectral image and its corresponding ground-truth depth map and it generates the simulated encoded sensor RGB image. The encoded image is then fed to the U-Net network, which produces the estimated depth. The loss between the estimated depth map and the ground truth depth is calculated to optimize the optical layer and network weights via back-propagation \cite{hecht1992theory}. In the next subsection, we present in more detail the forward model of the proposed system, including the parametrization of the CCA.

\subsection{Image Formation Model}

We create a camera model with a single convex thin lens with a focal length $f$ at distance $z_i$ from the sensor and a CCA in the lens as shown in Fig. \ref{fig:propagation}. Most of the real-world scenes contain objects at various depths that are imaged with different PSFs. To simulate the PSF for every depth $z$, consider a point emitter with wavelength $\lambda$ centered on the optical axis and located a distance $z$ away from the lens center. 

By propagating this point, the electric field of the spherical wave arriving at the lens is given by the Fresnel approximation\footnote{It assumes that the wavelength $\lambda$ is significantly smaller than the travel distance $z$: $\lambda \ll z$.} as:
\begin{equation}
U_{\text {in}}^{\lambda,z} (x,y) = \exp \left[ i \dfrac{2\pi}{\lambda} \dfrac{x^2+y^2}{z} \right].
\end{equation}
The wavefront then propagates through the lens with the lens phase delay being, $t(x, y)$: 
\begin{equation}
    t^\lambda(x, y)=\exp \left[-i \frac{k}{2 f}\left(x^{2}+y^{2}\right)\right],
\end{equation}
where $k=2 \pi / \lambda$ is the wavenumber. Because a lens has a finite aperture size, we insert an amplitude function $A(x, y)$ that blocks all light outside the open aperture. We also add the function $T(x,y,\lambda)$, which represents the CCA separated by a distance $z_t$ from the lens.\footnote{The proposed method assumes the CCA is attached to the lens, i.e. $z_t \approx 0$.} Assuming that the CCA wavelength response remains approximately constant over a spatial region of size $\Delta_m \times \Delta_m$ and wavelength interval $\Delta_d$, the CCA can be modeled using a rectangular function as follows: 
\begin{equation}
    T(x,y,\lambda) = \sum\limits_{i,j,\ell} {W_{i,j,\ell}}{\rm rect}\left({\frac{x}{{{\Delta _m}}} - i,\frac{y}{{{\Delta _m}}} - j,\frac{\lambda}{{{\Delta _d}}} - \ell} \right),
    \label{eq:ca_1}
\end{equation}
where $W_{i,j,\ell} \in [0,1]$ represents the discretized light response of the colored-coded aperture in the $(i,j)$-th position, and $\ell$-th spectral band of the discretized voxels \cite{bacca2021deep, arguello2021shift}. The above CCA model is generally applicable, where a square CCA with a filter matrix $N\times N$ and $L$ spectral bands results in $N^2L$ parameters $W_{i,j,\ell}$. However, practical constraints, such as manufacturing limitations \cite{arguello2014colored}, restrict the ability to customize filter spectral responses arbitrarily. In line with Henry et al. \cite{arguello2021shift}, the colors of the proposed CCA design are obtained by forming linear combinations of $4$ primary colors (green, red, blue, and cyan). This approach yields filter colors:
\begin{figure}[bt]
    \centering
    \includegraphics[width =0.8\columnwidth]{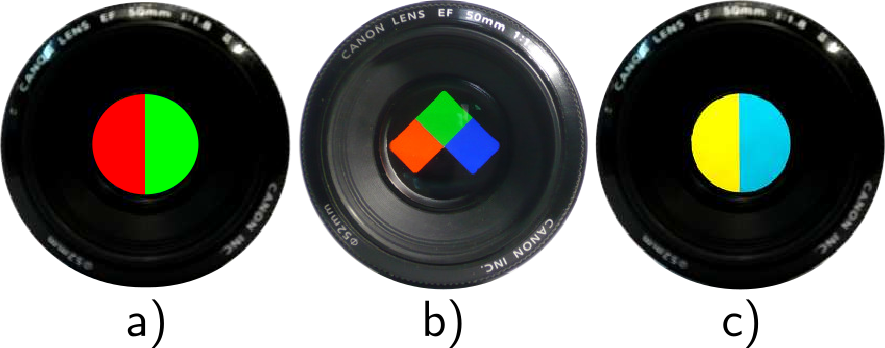}
    \caption{State-of-the-art color-coded apertures. a) \cite{amari1992single}, b) \cite{bando2008extracting} and c) \cite{paramonov2016depth}.}
    \label{fig:prev_cca}
\end{figure}
\begin{equation}
    {C^r}\!\left(\lambda \right) = \sum\limits_{\ell = 0}^{L - 1} \alpha _\ell^r {\rm rect}\left({\frac{\lambda}{{{\Delta _d}}} - \ell} \right),
    \label{eq:primar}
\end{equation}
where $\alpha_\ell^r$ represents the discrete spectral response of the $r$-th primary filter, which is predetermined based on available materials or chemical components. Using \eqref{eq:primar}, the CCA for our optical configuration can be finally expressed as:
\begin{equation}
\label{eq:CCA_}
    \begin{split}
    T\!\left({x,y,\lambda} \right) = \sum\limits_{i,j} \sum\limits_{r = 0}^{R - 1} \sum\limits_{\ell = 0}^{L - 1} {w_{i,j,r}} \\ \alpha _\ell^r
    {\rm rect}\left({\frac{x}{{{\Delta _m}}} - i,\frac{y}{{{\Delta _m}}} - j,\frac{\lambda}{{{\Delta _d}}} - \ell} \right),
    \end{split}
\end{equation}
where $R$ is the number of primary filters, and $w_{i,j,r}\in [0,1]$ represents the CCA weights for the $i$-th row, $j$-th column, and $r$-th primary filter. The electric field immediately after the lens and CCA is:
\begin{equation}
    U_{\text {out}}^{\lambda,z}(x, y)=A(x, y) T(x,y,\lambda) t^\lambda(x, y) U_{\text {in }}^{\lambda,z}(x, y).
    \label{Uout}
\end{equation}
We propagate $U_{\text {out}}^{\lambda,z}(x, y)$ a distance $z_i$ to the sensor using the exact transfer function \cite{goodman2005introduction}
\begin{equation}
    H_{{z_i}}\left(f_{x}, f_{y}\right)=\exp \left[i k z_i \sqrt{1-\left(\lambda f_{x}\right)^{2}-\left(\lambda f_{y}\right)^{2}}\right],
\end{equation}
to finally obtain the field in the sensor
\begin{equation}
    U_{\text {sen}}^{\lambda,z}\left(x^{\prime}, y^{\prime}\right)=\mathcal{F}^{-1}\left\{\mathcal{F}\left\{U_{\text {out }}^{\lambda,z}(x, y)\right\} \cdot H_{{z_i}}\left(f_{x}, f_{y}\right)\right\},
    \label{eq:Usen}
\end{equation}
where $\left(f_{x}, f_{y}\right)$ are spatial frequencies and $\mathcal{F}$ denotes the $2 \mathrm{D}$ Fourier transform. Since the sensor acquires light intensity, the PSF for each depth and wavelength is obtained using the magnitude-squared of \eqref{eq:Usen}: 
\begin{equation}
\label{eq:psf_SD}
\operatorname{PSF}_{\lambda, z}\left(x^{\prime}, y^{\prime}\right)=\left|U_{\text {sen} }^{\lambda,z}\left(x^{\prime}, y^{\prime}\right) \right|^{2}.
\end{equation}

\subsection{Depth-Dependent Image Formation Model}

The simulated PSFs are employed to approximate the captured image by an RGB sensor of a given 3D scene using a layered representation that models the scene as a set of surfaces on discrete depth planes \cite{hasinoff2007layer}. This allows the precomputation of a fixed number of PSFs corresponding to each depth plane. We make few modifications here to suit our datasets consisting of pairs of all-in-focus spectral images and their discretized depth maps. Considering an all-in-focus spectral image $\mathbf{I}$, and a set of $j= 1 \ldots J$ discrete depth layers, and corresponding occlusion masks $\mathbf{M}_j$, the encoded spectral image at wavelength $ \lambda$ is:
\begin{equation}
    \hat{\mathbf{I}}_{\lambda} = \sum _{j=1} ^ {J} (\mathbf{I_{\lambda}} * PSF_{\lambda,j}) \odot \mathbf{M}_{j},
    \label{eq:img_fomrm}
\end{equation}
where $*$ denotes 2D convolution and $\odot$ denotes element-wise multiplication. The cumulative occlusion masks $\mathbf{M}_j$ are alpha masks that modulate how much light from each layer is captured by the sensor at each pixel, to ensure smooth transitions between depths\cite{chang2019deep}. The alpha masks are generated by blurring binary depth masks $\mathbf{A}_j$ (indicating presence (1) or absence (0) of objects at the $j-$th depth layer) from current and preceding layers:
\begin{equation}
    \begin{array}{l}
\mathbf{M}_{j}^{\prime}=\left(1-\mathbf{M}_{j+1}\right)\left(\mathbf{A}_{j} * \operatorname{PSF}_{j}\right), for \quad j<J \\
\mathbf{M}^{\prime}{ }_{J}=\mathbf{A}_{J} * \operatorname{PSF}_{J}
\end{array}
\end{equation}
Here, layer $J$ is the layer closest to the camera that is not occluded by any additional layers, so $\mathbf{M}_{J}^{\prime}$ simply consists of the regions of the depth map that fall into layer $\mathbf{A}_{J},$ blurred by $\mathrm{PSF}_{J}$  \cite{chang2019deep}. Each layer behind $J$ is occluded by the layers in front of it. Finally, $\mathbf{M}_{j}^{\prime}$ are normalized to $\mathbf{M}_{j}$ such that the sum of occlusion mask weights at each pixel location sums to 1. Finally, the RGB channels of the captured image can be modeled as
\begin{equation}
    \ddot{\mathbf{I}}_{c}  = \int \hat{\mathbf{I}}_{\lambda} R_c(\lambda)) d\lambda, \text{ for $c$ = 1,2,3.}
\end{equation}
where $\mathcal{R}_c(\lambda)$ represents the spectral response of the $c$-th channel.

\subsection{Depth Reconstruction Network}

We use an encoder-decoder scheme based on convolutional neural networks, where the encoder is in charge of simulating the optical process of image acquisition, and the decoder extracts the depth information present in the encoded images. We adopt the U-Net architecture for the decoding process since it is widely used for pixel-level prediction, and it is adequate for depth estimation. This scheme is illustrated in Fig. \ref{fig:method}.

\subsubsection{Training Loss Function}

To optimize all the parameters of our model, the following cost functions were employed:
\begin{equation}
    \mathcal{L} = \alpha \mathcal{L}_{g} + \mu \mathcal{L}_{n} + \sigma \mathcal{L}_{s} % + \phi \mathcal{L}_{t},
\end{equation}
where $\mathcal{L}$ is the total loss and $\alpha, \mu, \sigma \in \mathbb{R}$ are weighting coefficients for each loss. The description of each term  of the total loss function is as follows:

\begin{itemize}[leftmargin=5mm]

\item \textbf{Gradient loss} $\mathcal{L}_{g}$: In an image, it is common to have different objects at different depths, which can generate sharp transitions in the estimated depth map. To account for this, we use an edge-dependent loss function defined given by:
\begin{equation}
\mathcal{L}_{g} = \frac{1}{M} \sum_{i,j} \| \nabla Y_{i,j} - \nabla \tilde{Y}_{i,j} \|^{2}
\end{equation}
where $\nabla$ is the 2-D Sobel edge detection operator \cite{sobel19683x3}, $Y \text{ and } \tilde{Y}$ represent the original and estimated depth maps, respectively, and $M$ is the total number of depth pixels. 

\item \textbf{Normal Loss} $\mathcal{L}_{n}$: In natural scenes, large objects have a greater impact on neural network depth estimation than small objects. This is because the network predicts depth by comparing it to surrounding object depths. To address this, we use a loss function, $\mathcal{L}_{n}$, which assesses the accuracy between the surface normal of the estimated depth map and the ground truth:
\begin{equation}
    \mathcal{L}_{n} = 1 - \dfrac{\left \langle n,\tilde{n} \right \rangle}{\max\left ( \left \| n \right \|_2, \left \| \tilde{n}  \right \|_2 \right  )},
\end{equation}
where $n = \left [ \nabla Y, 1 \right ]^T$ and $\tilde{n}  = \left [ \nabla \tilde{Y}, 1 \right ]^T$.

\item \textbf{Smoothness} $\mathcal{L}_{s}$: Because outliers may exist in some depth map estimates \cite{hu2019revisiting}, we chose a cost function that globally evaluates the estimated depth maps.
\begin{equation}
    \mathcal{L}_{s} = \begin{cases}0.5 d^{2} & \text { if }|d|<1 \\ |d|-0.5 & \text { otherwise, }\end{cases}
\end{equation}
where $ d = Y - \tilde{Y}$. $\mathcal{L}_{s}$ is a robust version of the $\ell_1$ norm that is more sensitive to outliers than the $\ell_2$ norm \cite{girshick2015fast}.

\end{itemize}

\subsubsection{CCA Constraint Design}

To design the CCA using the end-to-end approach, we use a composed constraint to ensure that the designed CCA is manufacturable. This constraint uses the CLIP function \cite{goodfellow2016deep} to restrict the values within the interval $[0,1]$ and normalization over primary filter colors coefficient ($w$ in \eqref{eq:CCA_}) to avoid the CCA amplifying the incident light in the lens. The constraint can be described as:
\begin{equation}
    w_{i,j,r} = \mathrm{CLIP}\left( \dfrac{w_{i,j,r}}{\sum_{r=0}^{R-1 }w_{i,j,r}}\right),
\end{equation}
where $R$ is the number of color filters in the CCA.

\subsubsection{Training details}

Given the forward model and the loss function, the chain rule can derive the back-propagation error. In our system, back-propagation is obtained using automatic differentiation implemented in the PyTorch framework \cite{paszke2019pytorch}. During the training, we use the Adam optimizer with parameters $\beta_{1}=0.99$ and $\beta_{2}=0.999$. Empirically, we found that using different learning rates for the optical layer and depth reconstruction network (U-Net) improves the performance. We used a training minibatch size of 32. The training and testing were performed on a NVIDIA GeForce RTX 3090 GPU.

\section{Simulations and Results}

\begin{figure*}[!t]
    \centering
    \includegraphics[width=0.99\textwidth]{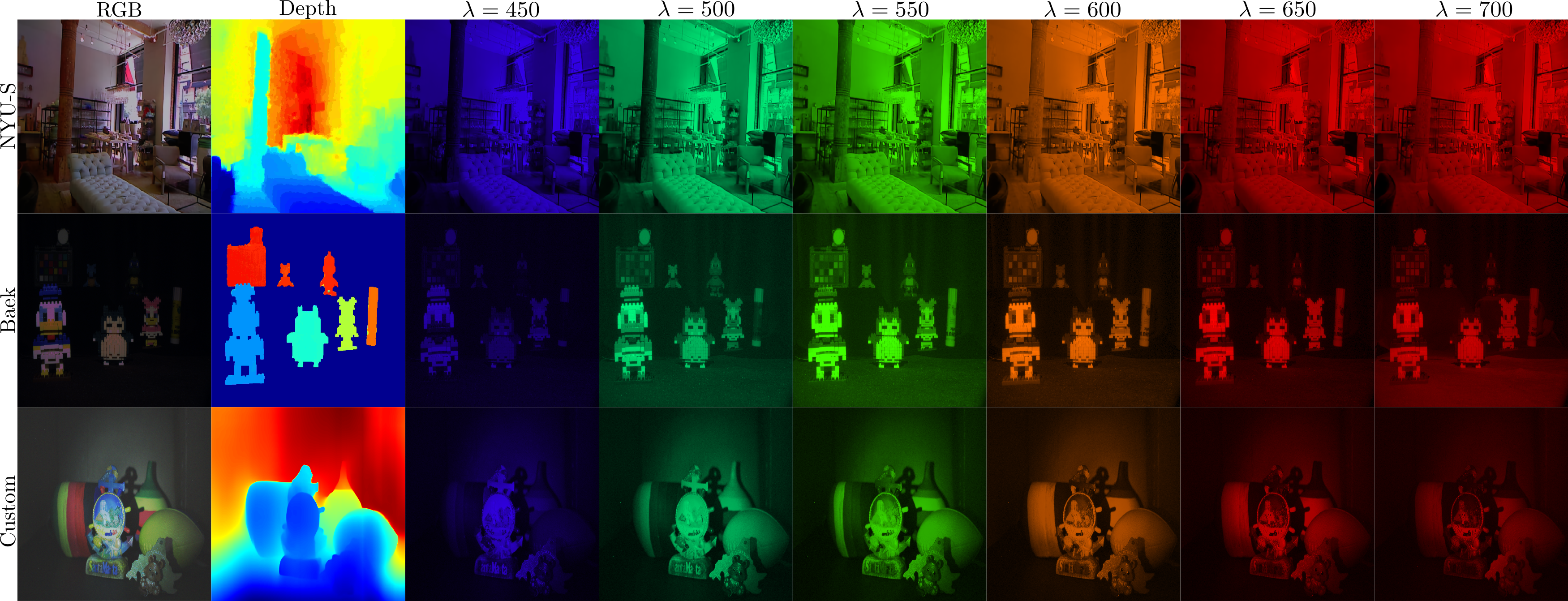}
    \caption{RGB image, depth map, and some spectral bands of scenes from the three datasets employed for validation. The RGB image shows the overall appearance of the scene, while the depth map provides information about the distance of objects from the camera. In this work, the intensities of different spectral bands are employed to codify depth information by leveraging a CCA.}
    \label{fig:DATA}
    \vspace{-3mm}
\end{figure*}

In this section, we evaluate the proposed approach in simulation on different datasets and provide comparisons with respect to other depth estimation methods. Furthermore, we performed ablation studies, which led us to discriminate the contribution of the principal components of our method.

\subsection{Datasets}

To perform the computational simulations, we need datasets that contain color or spectral information, therefore, the following datasets containing 31 spectral bands equally spaced from 400 to 700 nanometers were used:

\noindent $\bullet$ \textbf{NYU-S:} Since a large dataset with spectral and continuous depth information is not yet available, to prove the hypothesis of our approach, we generate fake spectral images with $31$ bands from the RGB images in the NYU dataset \cite{Silberman:ECCV12} by using the CNN proposed by Yuzhi et al. \cite{zhao2020hierarchical}, see Fig. \ref{fig:DATA}. The NYU-S dataset contains 1450 fake spectral images with their corresponding depth maps. Additionally, we used the default training/testing splits provided in \cite{Silberman:ECCV12} and performed random cropping with window sizes of $256 \times 256$ to augment the training set.

\noindent $\bullet$ \textbf{Baek:} Real spectral-depth dataset published in Baek et al. \cite{baek2021single} (see Fig. \ref{fig:DATA}). This dataset contains 18 spectral images with $31$ bands and their corresponding depth map. During training, 13 images were used and 5 images were used for testing. To increase the variability of the data during training, a random cropping with window sizes of $512 \times 512$ was conducted.

\begin{figure}[!b]
    \centering
    \includegraphics[width=0.9\columnwidth]{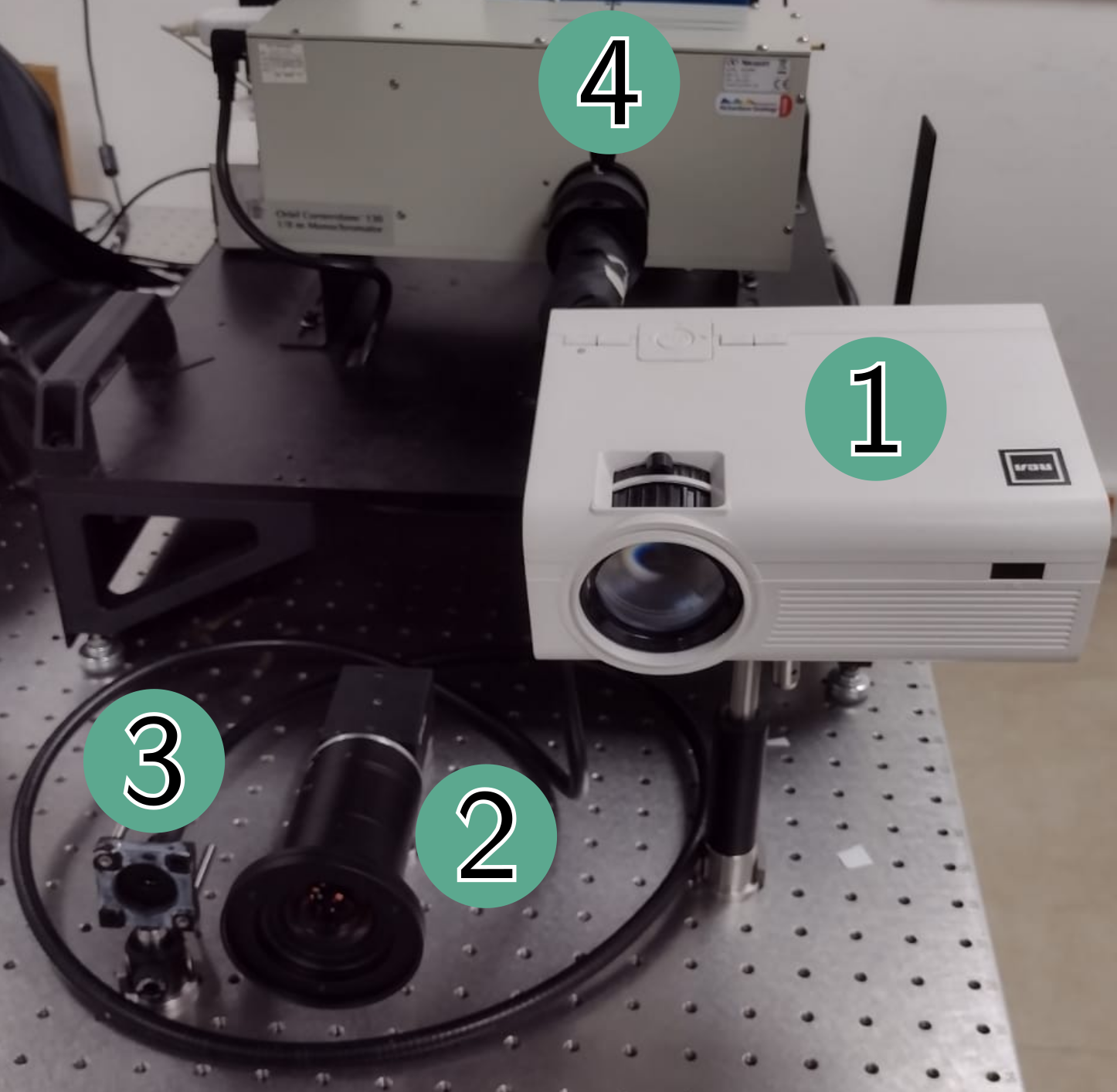}
    \caption{Optical system used to acquire our own spectral-depth dataset using spectral scanning and structured light. The light projector (1) emits a collimated beam of structured light patterns that is captured by the camera (2). The tunable light source (4) emits selective wavelengths to travel through an optical fiber (3) to illuminate the scene which is also captured with camera (2).}
    \label{fig:Opt_arch}
\end{figure}

\noindent $\bullet$ \textbf{Custom}: Due to the limited amount of spectral-depth data from the dataset mentioned above, we decided to acquire our own dataset, which contains 63 spectral images with their corresponding depth map, see Fig. \ref{fig:DATA}. The acquisition of the spectral images was carried out by illuminating the scene with a tunable light source \footnote{https://www.newport.com/p/TLS130B-300X} (TLS), which offers wavelengths in the range of $300-1800$ nm. We used wavelengths between $400-700$ nm with intervals of $10$ nm. The depth maps vary in a range of $0.4-1.6$ m, estimated using the structured light (SL) setup shown in Fig.\ref{fig:Opt_arch} and the phase unwrapping via graph cuts algorithm \cite{bioucas2007phase}. 

\subsection{Evaluation}

Following previous works \cite{chang2019deep, ranftl2021vision}, we adopt the following evaluation metrics to quantitatively assess the performance of our depth prediction model. Specifically, we use: 

\noindent $\bullet$ Mean Absolute Error (MAE):  
    $\frac{1}{N}\sum_{i=1}^N |y_i - \tilde{y}_i|$
    
\noindent $\bullet$ Mean Relative Error (REL): 
    $\frac{1}{N} \sum_{i} ^ N \frac{\left|y_{i}-\tilde{y}_{i}\right|}  {\tilde{y}_{i}} $
    
\noindent $\bullet$ Mean Log$_{10}$ Error (Log$_{10}$): 
    $\frac{1}{N} \sum_{i} ^ N \left|\log_{10}(y_{i})- \log_{10}(\tilde{y}_{i})\right|$
    
\noindent $\bullet$ Root Mean Squared (RMSE): 
    $\sqrt{\frac{1}{N} \sum_{i} ^ {N} \left|y_{i}-\tilde{y}_{i}\right|^{2}} $
    
\noindent $\bullet$ Thresholded Accuracy ($\delta_j$): $\frac{1}{N} \sum_{i}^N \max\left(\frac{\tilde{y}_{i}}{y_{i}},\frac{y_{i}} {\tilde{y}_{i}}\right)< 1.25^j$,
where $y_{i}$ and $\tilde{y}_{i}$ are ground-truth and estimated depth map respectively, $N$ is the total depth map pixels and $j=\left \{ 1,2,3 \right \} $. Smaller values on REL, Log$_{10}$, and RMS error are better, and higher values on $\delta_j$ threshold are better.

\subsection{CCA design}
\label{sec:cca_design}

\begin{table}[!b]
\vspace{-7mm}
\caption{Quantitative results of ablation studies on the NYU-S dataset. The best results are in bold, and the second-best are underlined.}
\resizebox{\columnwidth}{!}{%
\begin{tabular}{|c|cccccc|}
\hline
\textbf{Exp} & \textbf{RMS}$\boldsymbol{\downarrow}$   & \textbf{REL}$\boldsymbol{\downarrow}$   & $\text{\textbf{Log}}_{10}\boldsymbol{\downarrow}$ & $\delta_1 \boldsymbol{\uparrow}$    & $\delta_2 \boldsymbol{\uparrow}$     & $\delta_3 \boldsymbol{\uparrow}$ \\ \hline
Vanilla       & 0.848          & 0.276          & \multicolumn{1}{c|}{0.107}          & 0.619          & 0.856          & 0.942          \\
Fixed-BCA       & 0.226          & 0.055          & \multicolumn{1}{c|}{0.024}          & 0.975          & 0.996          & 0.994          \\
Levin-BCA:      & 0.208          & 0.053          & \multicolumn{1}{c|}{0.022}          & {\ul 0.979}    & {\ul 0.998}    & {\ul 0.998}    \\
Learned-BCA         & {\ul 0.196}    & {\ul 0.051}    & \multicolumn{1}{c|}{{\ul 0.021}}    & {\ul 0.979}    & 0.996          & {\ul 0.998}    \\
Bando-CCA     & 0.224          & 0.067          & \multicolumn{1}{c|}{0.028}          & 0.968          & 0.995          & 0.996          \\
Fixed-CCA      & 0.214          & 0.060          & \multicolumn{1}{c|}{0.025}          & 0.968          & 0.996          & 0.997          \\
Learned-CCA    & \textbf{0.147} & \textbf{0.039} & \multicolumn{1}{c|}{\textbf{0.016}} & \textbf{0.993} & \textbf{0.998} & \textbf{0.999} \\ \hline
\end{tabular}%
}
\label{tab:ablation}
\end{table}

We compared the proposed depth estimation approach with several baseline approaches based on CAs and CCAs to demonstrate the advantages of the proposed CCA design. The baselines were trained using the NYU-S dataset and are as follows:

\begin{itemize}
    \item \textbf{Vanilla:} Training of the vanilla U-Net neural network for depth estimation, without the addition of optical layers and a learning rate of $5e^{-4}$.
    \item \textbf{Fixed-BCA:} In this set-up, we fixed a random binary coded aperture (BCA), which filters the light entirely in certain regions of the image plane, and only optimizes a U-Net for depth estimation.
    \item \textbf{Levin-BCA \cite{levin2007image}} This baseline is similar to the previous one, with the difference that the BCA is the code optimized with genetic algorithms by Levin et al. \cite{levin2007image} for depth estimation.
    \item \textbf{Learned-BCA:} This baseline uses the proposed end-to-end scheme to obtain an optimal BCA for monocular depth estimation, using a learning rate of $1^{-3}$ for the optical layer and a learning rate of $5^{-4}$ for the neuronal network. We use the optimization scheme proposed by Bacca et al. \cite{bacca2021deep} to obtain binary values. 
    \item \textbf{Bando-CCA \cite{bando2008extracting}:} In this baseline, the RGB-coded aperture proposed by Bando et al. \cite{bando2008extracting} is used, and optimize the U-Net parameters to extract the depth map from the encoded image, with a learning rate of $5e^{-4}$.
    \item \textbf{Fixed-CCA:} In this set-up, we fixed a random CCA and trained only U-Net parameters with a learning rate of $5^{-4}$.
    \item \textbf{Learned-CCA:}  For our proposed approach, the learning rates for the optical layer and U-Net were $5e^{-2}$ and $5e^{-4}$, respectively.
\end{itemize}

Table \ref{tab:ablation} presents the quantitative results of the different baselines compared with the proposed approach. The \textit{Vanilla} approach demonstrates that using only the U-Net architecture without incorporating optical coding is inadequate for depth estimation, as all other experiments using a coded aperture achieve superior performance. In particular, the \textit{Levin-BCA} and \textit{Learned-BCA} baselines reveal that using an optimized coded aperture contributes to more accurate depth estimation compared to the \textit{Fixed-BCA} baseline. Comparing the colored approach \textit{Bando-CCA} with the naive \textit{Fixed-CCA} approach demonstrates that using more colors improves depth estimation. However, \textit{Bando-CCA} and \textit{Fixed-CCA} baselines also show that the use of CCA without any optimization criteria produces suboptimal results, as is evident from the superior performance obtained in the \textit{Levin-CA} and \textit{Learned-BCA} experiments. The proposed approach achieves the best performance in all metrics highlighting the advantage of the learned color codification for high-throughput depth estimation. 

\begin{table}[!t]
\caption{Quantitative comparison of the proposed approach with state-of-the-art models on the NYU-Depth test set. The best results are in bold, and the second-best are underlined.}
\centering
\resizebox{\columnwidth}{!}{%
\begin{tabular}{|c|ccc|ccc|}
\hline
\textbf{Model} & \textbf{REL}$\boldsymbol{\downarrow}$   & \textbf{RMS}$\boldsymbol{\downarrow}$   & $\text{\textbf{Log}}_{10}\boldsymbol{\downarrow}$ & $\delta_1 \boldsymbol{\uparrow}$    & $\delta_2 \boldsymbol{\uparrow}$     & $\delta_3 \boldsymbol{\uparrow}$     \\ \hline
DORN \cite{fu2018deep}             & 0.115          & 0.509          & 0.051          & 0.828          & 0.965          & 0.992          \\
MS-CRF  \cite{xu2017multi}         & 0.121          & 0.586          & 0.052          & 0.811          & 0.954          & 0.987          \\
2S-CNN \cite{li2017two}            & 0.152          & 0.611          & 0.064          & 0.789          & 0.955          & 0.988          \\
Make3D \cite{saxena2007learning}   & 0.349          & 1.214          & -              & 0.447          & 0.745          & 0.897          \\
UP-CNN \cite{laina2016deeper}      & 0.127          & 0.573          & 0.055          & 0.811          & 0.953          & 0.988          \\
DPT \cite{ranftl2021vision}        & 0.110          & 0.357          & 0.045          & 0.904          & 0.988          & {\ul 0.998}    \\
VNL \cite{yin2019enforcing}        & 0.108          & 0.416          & 0.048          & 0.875          & 0.976          & 0.994          \\
BTS \cite{lee2019big}              & 0.110          & 0.392          & 0.047          & 0.885          & 0.978          & 0.995          \\
LapDepth \cite{song2021monocular}  & 0.105          & 0.384          & 0.045          & 0.895          & 0.983          & 0.996          \\
AdaBins \cite{bhat2021adabins}     & 0.103          & 0.364          & 0.044          & 0.903          & 0.984          & 0.997          \\ \hline
Chang et al. \cite{chang2019deep}      & 0.087          & 0.432          & 0.052          & 0.930          & 0.990          & {\ul 0.998}    \\
Wu et al. \cite{wu2019phasecam3d} & 0.093          & 0.382          & 0.050          & 0.932          & 0.989          & 0.997          \\
Learned-BCA                             & {\ul 0.051}    & {\ul 0.196}    & {\ul 0.021}    & {\ul 0.979}    & {\ul 0.996}    & {\ul 0.998}    \\
Learned-CCA                            & \textbf{0.039} & \textbf{0.147} & \textbf{0.016} & \textbf{0.993} & \textbf{0.998} & \textbf{0.999} \\ \hline
\end{tabular}%
}
\label{tab:state-art}
\end{table}

\subsection{Monocular depth estimation}

\begin{figure*}[!t]
    \centering
    \includegraphics[width=\textwidth]{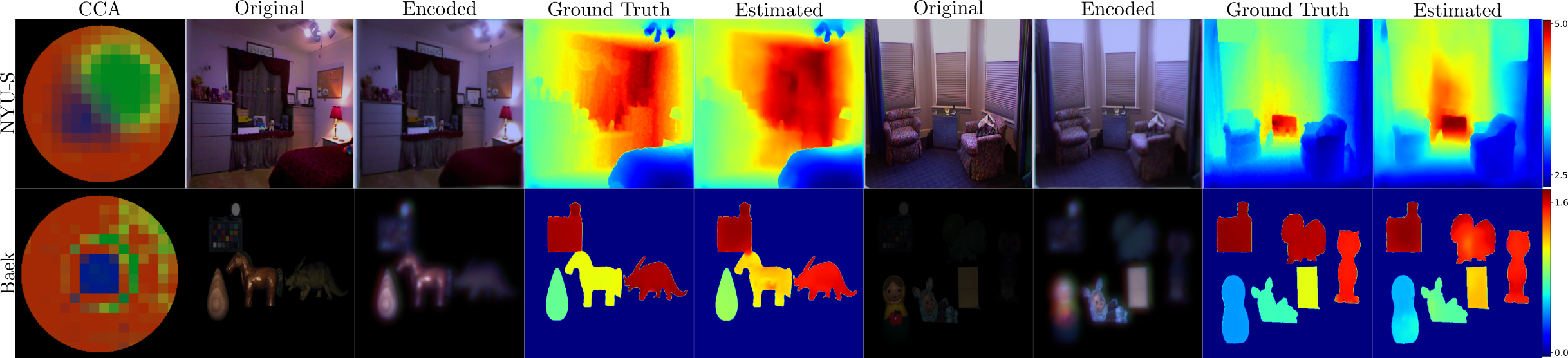}
    \caption{Visual results on the \textbf{NYU-S} and \textbf{Beak} datasets, along the learned color-coded aperture. The reconstructed depth maps closely match the ground truth depth maps in both datasets. %The color-coded apertures show the relative importance of color in the retrieval depth process.
    }
    \label{Fig:Results}
\end{figure*}
Once validated our proposed CCA-based depth estimation approach, we compare our method with state-of-the-art monocular approaches \cite{ranftl2021vision,fu2018deep,xu2017multi,li2017two,saxena2007learning,laina2016deeper,yin2019enforcing,lee2019big,song2021monocular,bhat2021adabins} that employ all-in-focus RGB images (without optimized optical coding). We also compare our method with approaches that perform end-to-end optimization of DOEs that promote chromatic aberrations as depth cues \cite{chang2019deep, wu2019phasecam3d,baek2021single}. Additionally, we include the results of the proposed approach but the coded aperture is constrained to be binary (\textit{Learned-BCA}). The quantitative results performed in the default NYU-S test set are compiled in Table \ref{tab:state-art}. Our colored-controlled-defocus approach achieves superior accuracy in all metrics, demonstrating that coding across the visible spectrum contributes to improved depth estimation using a simple decoder such as the U-Net. Furthermore, the experimental results show that optimizing BCA using the proposed approach achieves comparable performance to state-of-the-art DOEs. The top of Fig. \ref{Fig:Results} shows the learned CCA, examples of encoded images, and estimated depth maps for NYU-S datasets. These qualitative results show that the estimated depth maps preserve high fidelity to the ground truth. 
\begin{table}[!t]
\caption{Quantitative results over test-set in the Baek dataset.}
\centering
\resizebox{\columnwidth}{!}{%
\begin{tabular}{|c|c|c|c|c|}
\hline
Method & Chang et al. \cite{chang2019deep} & Wu et al. \cite{wu2019phasecam3d} & Baek et al. \cite{baek2021single}  & Ours          \\ \hline
RMSE   & 0.45         & 0.31      & 0.20          & \textbf{0.18} \\
MAE    & 0.25         & 0.19      & \textbf{0.12} & \textbf{0.12} \\ \hline
\end{tabular}%
}
\label{Tab:Hayato}
\end{table}
\\ \\
\noindent \textbf{Performance on Baek dataset}
Finally, we learned our encoder-decoder model using Baek dataset. Visual results of the learned CCA, encoded image, and recovered depth maps are shown at the bottom of Fig. \ref{Fig:Results}, demonstrating also accurate depth estimation. Since the Baek dataset contains few images with considerable dark areas, there is a noticeable difference in the obtained CCA in this experiment compared to the CCA learned with the NYU-S dataset. Additionally, Table \ref{Tab:Hayato} presents quantitative results (adapted from \cite{baek2021single}) comparing our method with state-of-the-art approaches that learn DOEs \cite{chang2019deep, wu2019phasecam3d,baek2021single}. They show that our method achieves similar or higher results with the advantage of using an optical element that can be implemented with a low-cost photographic film as demonstrated in the following section.

\section{Experiments on Real Hardware}

\begin{figure}
    \centering
    \includegraphics[width=\columnwidth]{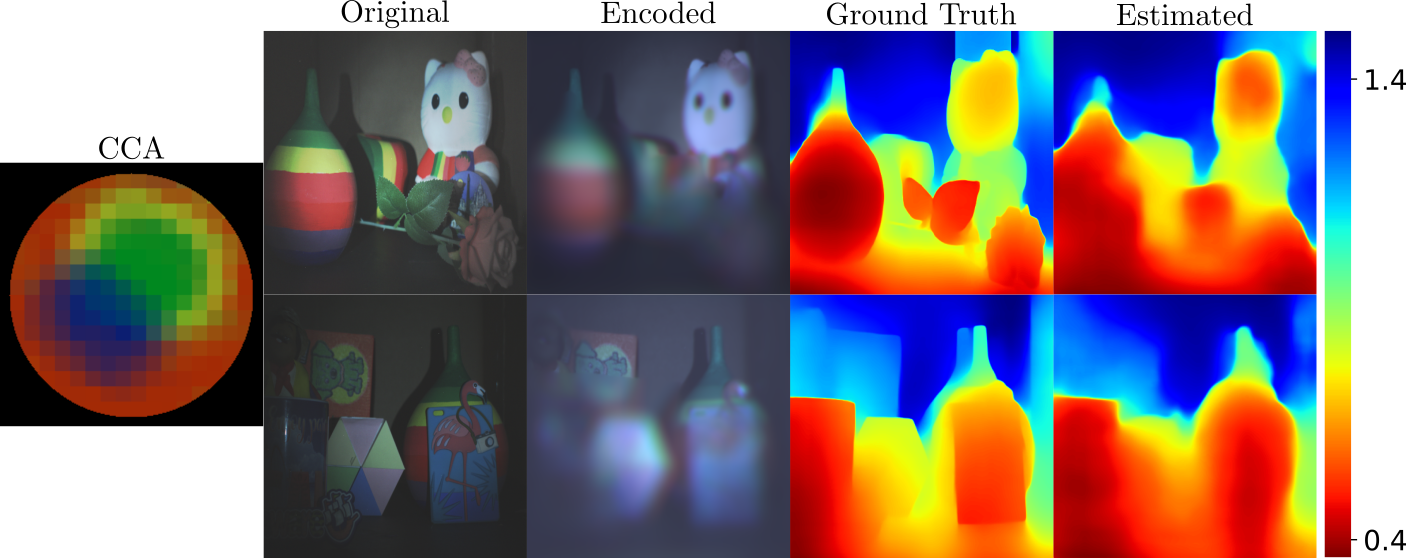}
    \caption{Visual results using the Custom dataset, with the optimized color-coded aperture. The reconstructed depth maps closely match the ground truth depth maps.}
    \label{Fig:Results_Custom}
\end{figure}

After rigorously validating our method with state-of-the-art datasets in simulation, we validate our proposed approach with real captures. First, the proposed model is learned using our \textit{Custom} dataset. Figure \ref{Fig:Results_Custom} shows optimized CCA, encoded images, and corresponding estimated depth maps. The depth maps largely match the actual values, demonstrating the robustness of our method. Note that the learned CAs using NYU-S and \textit{Custom} data have similar structures; confirming that the interpolated spectral images in NYU-S yield close results to spectral images acquired with a real spectral imager.

\begin{figure}[h!]
    \centering
    \includegraphics[width=\columnwidth]{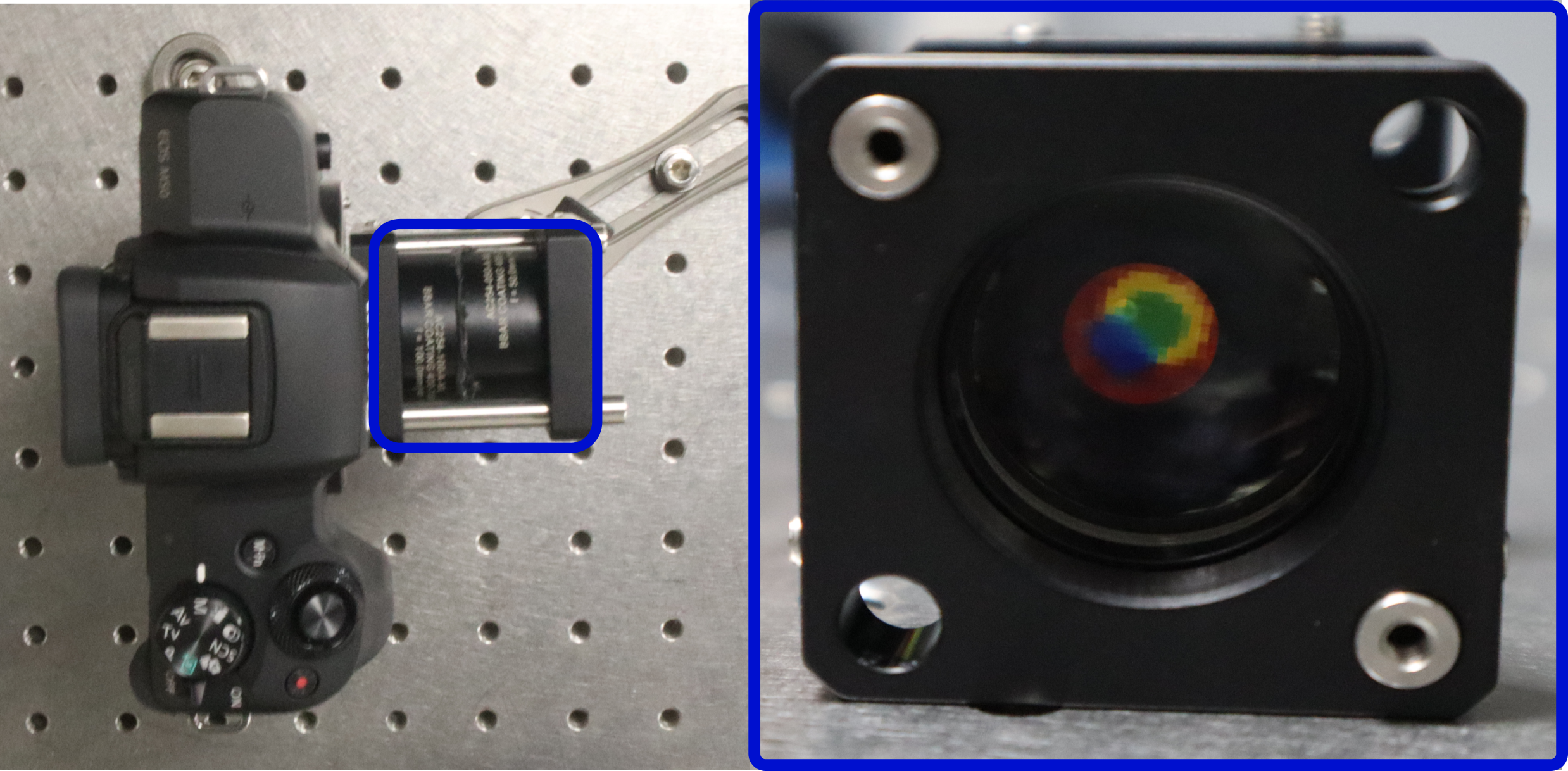}
    \caption{Testbed implementation of the proposed optical system along the optimized CCA implemented with a color photographic film.}
    \label{fig:Lab_mount}
\end{figure}
\begin{figure}[h!]
    \centering
    \includegraphics[width=\columnwidth]{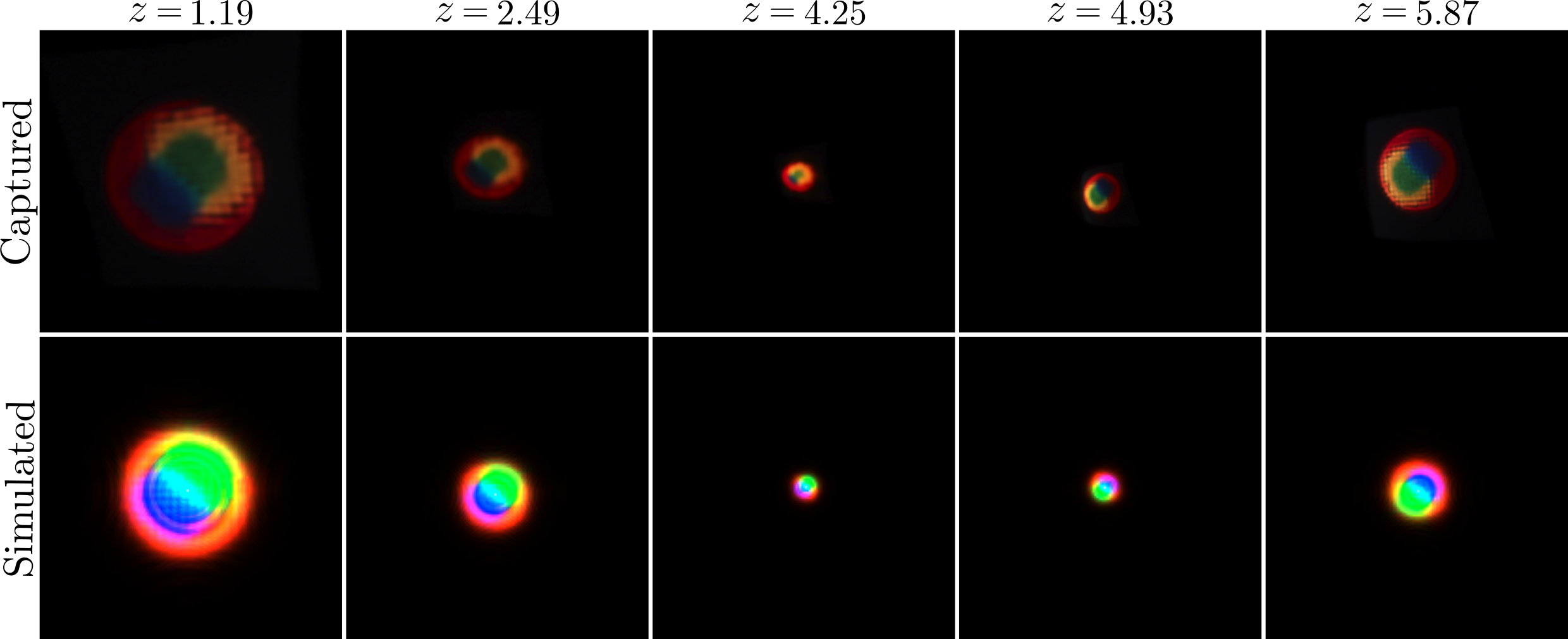}
    \caption{Real and simulated PSFs at different depths in meters.}
    \label{fig:PSFs_lab}
\end{figure}

%After rigorously validating our method with state-of-the-art datasets in simulation, we validate our proposed approach with real captures acquired with our own built prototype.
Second, we build a test-bed prototype shown in Fig. \ref{fig:Lab_mount}. This system incorporates two Thorlabs Achromatic Doublets lenses \footnote{\url{https://www.thorlabs.com/newgrouppage9.cfm?objectgroup_id=2696}} with focal lengths of 50 mm and 100 mm. These lenses are carefully arranged to form a composite lens configuration, with our optimized CCA positioned at the center. This configuration aims to closely approximate the simulated model, as depicted in Fig. \ref{fig:propagation}. This system is adapted to a Canon camera (EOS M50) with a CMOS sensor of $24.1$ megapixels (see Fig. \ref{fig:Lab_mount}). To fabricate the optimized CCA, we use a color FUJICHROME Velvia $50$ transparency film \cite{b&amp} ($35$ mm, $36$ exposures) and the printing process used in \cite{arguello2021shift}. The printed CCA is shown in the inset of Fig. \ref{fig:Lab_mount}. Once the system is implemented, the performance of the proposed method may be affected by mismatch and fabrication errors of the CCA. Therefore, we capture the real PSFs (see Fig \ref{fig:PSFs_lab}) by using a white light point source and calibrating the simulated PSFs with the captured PSFs. Then, with the calibrated PSFs, we fine-tune the U-Net depth estimation network for $100$ epochs, a learning rate of $3e-5$, and our \textit{Custom} dataset.

\begin{figure}[!t]
    \centering
    \includegraphics[width=\columnwidth]{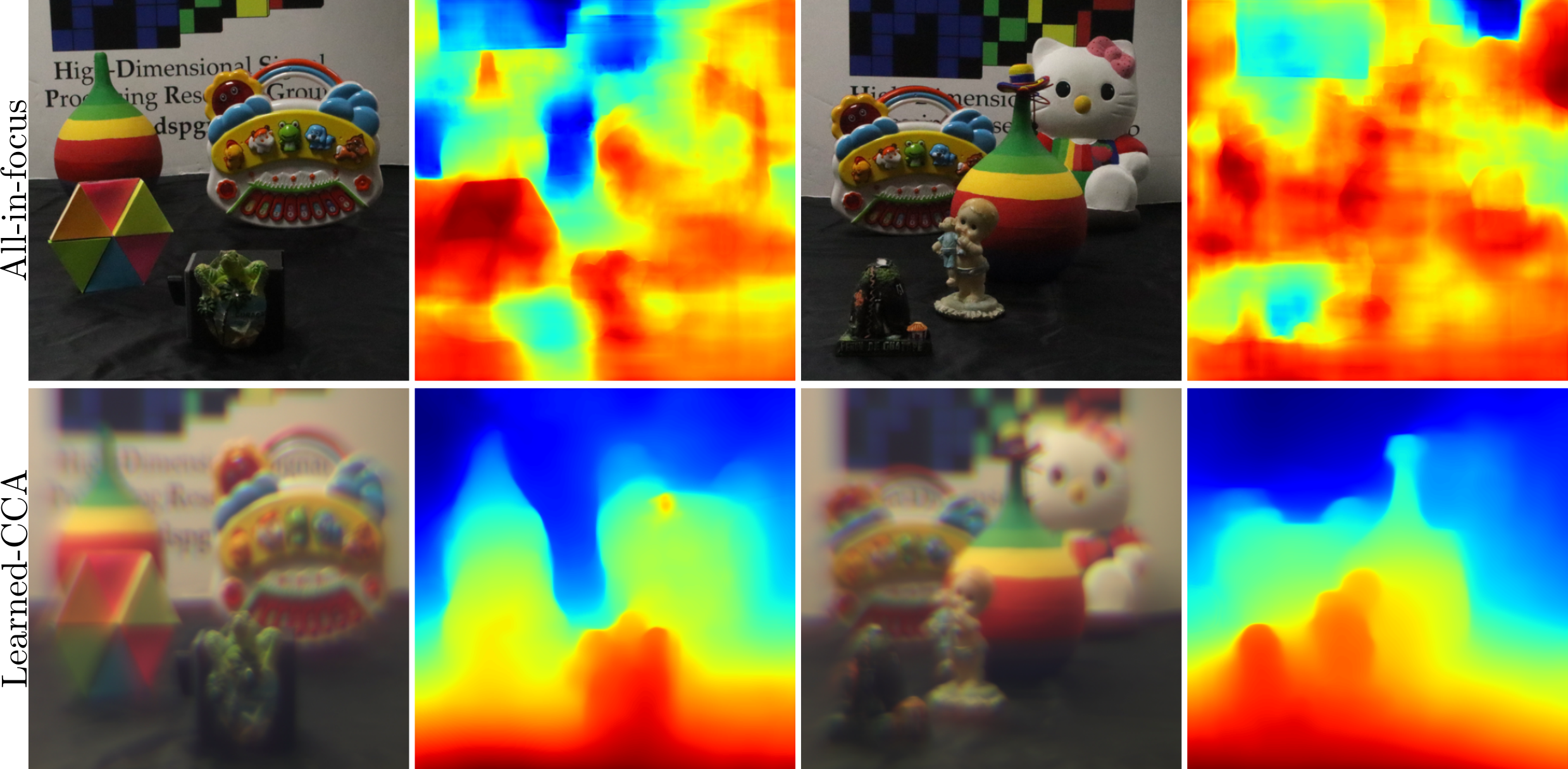}
    \caption{Real depth map predictions from our proposed spectral encoded images and all-in focus images.}
    \label{fig:Custom_results}
\end{figure}

We also capture an all-in-focus version of the same scene using an identical camera setup without employing any optical coding and estimate a depth map utilizing a U-Net trained using the NYU dataset. The real captured images and depth map estimates are illustrated in Fig. \ref{fig:Custom_results}. These results, demonstrate that our proposed approach (\textit{Learned-CCA}) exhibits remarkable robustness in accurately estimating depth maps within indoor scenes, in contrast to the approach without any coding methodology. Furthermore, the results of this experiment demonstrate the feasibility of implementing the proposed spectral codification by leveraging color photographic films.

\section{Conclusions}

In this work, we design a color-coded aperture (CCA) using an end-to-end optimization approach for encoding the depth information using a monocular system. Compared with CCA-based approaches that employ few colors for encoding depth information, the proposed approach demonstrates that using more color filters with rich spectral information significantly improves depth estimation. On the other hand, our approach ratifies, that the data-driven design of specific-purpose optical elements using the E2E technique, allows better depth encoding, compared to CCA used without any data constraint. We built a prototype camera using the optimized CCA and recover high-fidelity depth maps using a fine-tuned neural network validating our proposed approach in real scenarios. The proposed color codification can be fabricated using a low-cost photographic film and established film development, highlighting the additional economic benefit of the proposed codification compared with high-cost DOEs.



\vfill

\end{document}